\documentclass[a4paper,twoside,10pt]{article}
\usepackage[T1]{fontenc}
\usepackage[latin1]{inputenc}
\usepackage{amsmath, amssymb, latexsym, amsthm, mathrsfs}
\usepackage{graphicx}
\usepackage{ulem}
\usepackage{dsfont}
\input{epsf}
\usepackage{hyperref}

\date{}
\def\XXint#1#2#3{{\setbox0=\hbox{$#1{#2#3}{\int}$}
\vcenter{\hbox{$#2#3$}}\kern-.5\wd0}}

\renewcommand{\S}{\mathbb{S}}
\renewcommand{\epsilon}{\varepsilon}
\renewcommand{\phi}{\varphi}

\DeclareMathOperator*{\argmin}{argmin}
\DeclareMathOperator*{\diag}{diag}
\DeclareMathOperator*{\logsumexp}{logsumexp}

\theoremstyle{plain}

\theoremstyle{remark}

\title{Implementation of batched Sinkhorn iterations for entropy-regularized Wasserstein loss}
\author{Thomas Viehmann
\thanks{MathInf GmbH, tv@mathinf.eu}
}
\begin{document}

\maketitle

\begin{abstract}
  In this report, we review the calculation of entropy-regularised Wasserstein loss introduced by Cuturi and document a practical implementation in PyTorch.
\end{abstract}

Recently the Wasserstein distance has seen new applications in machine learning and deep learning. It commonly replaces the Kullback-Leibler divergence (also often dubbed cross-entropy loss in the Deep Learning context). In contrast to the latter, Wasserstein distances not only consider the values probability distribution or density at any given point, but also incorporating spatial information in terms of the underlying metric regarding these differences. Intuitively, it yields a smaller distance if probability mass moved to a nearby point or region and a larger distance if probability mass moved far away.

There are two predominant variants of Wasserstein distance approximations used in machine learning:
\begin{itemize}
\item Stochastically optimised online estimates of the Wasserstein distance. This is the concept underpinning many of the GAN applications using a (heuristic approximation of) the Wasserstein distance as a \textit{discriminator}. Starting from the Wasserstein GAN \cite{WGAN} as an improvement over the KL-based DCGAN, with improvements to how to estimate the Wasserstein distance in WGAN-GP \cite{WGAN-GP}, and SN-GAN \cite{SN-GAN}.
\item Direct computation of the Wasserstein distance as a replacement for the cross-entropy loss in mini-batch training. This is commonly done using the entropy regularised Wasserstein distance and the Sinkhorn iterations \cite{Cuturi}. In the context of deep learning this has been proposed by \cite{Frogner}, but there is also earlier work in image retrieval using the (non-regularised) Wasserstein distance, see e.g. \cite{EMD-ImageRetrieval}. A comprehensive overview is given in \cite{PeyreCuturi}.
\end{itemize}
In the present note we will be concerned with this latter use of the Wasserstein distance.
One of the challenges is the numerical stability of the Sinkhorn iteration and carrying that over to mini-batch computations efficiently. We propose an enhanced method in the combination of Sections~\ref{sec_backward} and \ref{sec_stable}. While the ingredients appear to be  readily available, it seems that they have not been put together in recent implementations we observed.

\section{A brief review of the Wasserstein distance and its entropy regularisation}
\label{sec_review}

We review the regularised Wasserstein distance and focus on probability distributions on finite sets. We largely follow \cite{Cuturi}.

For positive integers $d_1$ and $d_2$, consider two probability measures $\mu \in \mathbb{R}^{d_1}$ and $\nu \in \mathbb{R}^{d_2}$ on the set of $d_1$ and $d_2$ points, i.e. $\mu_i , \nu_j \geq 0$ and $\sum_i \mu_i = \sum_j \nu_j = 1$.

A coupling $P \in U := \{ P \in \mathbb{R}^{d_1 \times d_2} | p_{ij} \geq 0, \quad \sum_j p_{ij} = \mu_i, \quad \sum_i p_{ij} = \nu_i  \}$ of $\mu$ and $\nu$ is a probability measure with marginals $\mu$ and $\nu$. Intuitively, a coupling can be interpreted as a mapping of the probability mass of $\mu$ to that of $\nu$.

We introduce a cost on the set of couplings by means of a $d_1 \times d_2$-matrix  $c_{ij} \geq 0$. Then a coupling $P$ between $\mu$ and $\nu$ is optimal for $C$ if it is a minimiser of $E^0(P) = \sum_{ij} p_{ij} c_{ij}$ and $U$. By compactness of the admissible set, such a minimiser always exists, but in general it is not unique.

Cuturi \cite{Cuturi} proposed to consider the regularised functional
\[
  E^\lambda(P) = \sum_{ij} p_{ij} c_{ij}  - \lambda h(P),
\]
where $h(P)$ is the entropy $h(P) = - \sum_{ij} p_{ij} \log p_{ij}$.

As \cite{Cuturi} notes, the minimisation problem is closely related to the problem of minimising the original functional on a restricted set with entropy $h(P) \geq \alpha$: If a minimiser $P^*$ of  $E^\lambda(P)$ has entropy $h(P^*) = \alpha$, then it is a minimiser of the original functional $E^0$ on $U_\alpha := U \subset \{ h(P) \geq \alpha \}$: Another coupling $P' \in U_\alpha$ has $E^0(P') = E^\lambda(P') + h(P') \geq E^\lambda(P^*) + \alpha = E^0(P^*)$, where we use the that $P^*$ is an $E^\lambda$-minimiser and the admissibility condition for $P' \in U_\alpha$.

Note that a probability distribution becomes more ``regular'' with increasing entropy, motivating the lower bound and negative sign for the entropy term.

As the entropy is strictly concave and the cost term is linear in $p_{ij}$, the functional $E^\lambda(P)$ is convex and has a unique minimum on the admissible set $U$.

To characterise the minimum we introduce the Lagrange multipliers $\alpha \in \mathbb{R}^{d_1}$ and $\beta \in \mathbb{R}^{d_2}$ to capture the equality constraints in the definition of $U$ and have the augmented functional
\[
  E^\lambda_{\mu, \nu}(P, \alpha, \beta) = E^\lambda(P)
        + \sum_i \alpha_i (\sum_{j} p_{ij} - \mu_i)
        + \sum_j \beta_j (\sum_{i} p_{ij} - \nu_j)
\]
and minimax problem
\[
  P^* = \argmin_{P \in U} \sup_{\alpha \in \mathbb{R}^{d_1}, \beta \in \mathbb{R}^{d_2}} E^\lambda_{\mu, \nu}(P, \alpha, \beta)
\]

We write the Euler-Lagrange equations
\begin{equation}
  \label{eq_elg}
  0 = \frac{\partial}{\partial p_{ij}} E^\lambda_{\mu, \nu}(P, \alpha, \beta)
    = c_{ij} - \lambda - \lambda \log p_{ij} + \alpha_i + \beta_j.
\end{equation}
Solving for $p_{ij}$ we get
\begin{equation}
  \label{eq_pij_form}
  p_{ij} = \exp(-1-\frac{1}{\lambda} \alpha_i-\frac{1}{\lambda} \beta_j - \frac{1}{\lambda} c_{ij}) 
\end{equation}
for appropriate $\alpha$ and $\beta$. We absorb $\alpha$, $\beta$ and the constant by introducing $u := \exp(-\frac{1}{2}\mathbf{1}-\frac{1}{\lambda} \alpha)$ and $v := \exp(-\frac{1}{2}\mathbf{1}-\frac{1}{\lambda} \beta)$. We also write $K := \exp(-\frac{1}{\lambda} M)$

With this equation \eqref{eq_pij_form} becomes
\[
 P = \diag(u) K \diag(v).
\]
where $\exp$ is the element wise exponential function and $\diag$ the diagonal embedding operator mapping vectors to diagonal matrices.
Plugging this representation into the marginal constraints $\mu = P \mathbf{1}$ and $\nu = P^T \mathbf{1}$ and we get the coordinate-wise equations
\[
  \mu_i = u_i (Kv)_i , \qquad
  \nu_j = v_j  (K^T u)_j,
 \]
and solving for $u$, $v$ we have
\[
  \mu_i / (Kv)_i = u_i  , \qquad
  \nu_j / (K^T u)_j = v_j  .
\]
This makes is natural to set up the celebrated Sinkhorn-Knopp iteration
\begin{equation}
  \label{eq_sk_plain}
  v^{(k+1)}_j := \nu_j / (K^T u^{(k)})_j  , \qquad
   u^{(k+1)}_i :=  \mu_i / (Kv^{(k+1)})_i 
\end{equation}

This iteration alternatingly enforces each of the two marginal constraints.
An important algorithmic question that we skip here is the convergence of this fixed-point iteration, see e.g. \cite{SKConv}.

\section{Derivative}
\label{sec_backward}

Recall that the Lagrange multiplier gives the derivative of $E^\lambda$ with respect to the constraint. However, as s $\mu$ and $\nu$ are themselves probability measures, they are constrained themselves: To preserve their summing to $1$, the allowed variations are only those of mean $0$.

Thus, to compute a meaningful gradient with respect to the input manifold, we have to project the full gradient $\alpha$ by subtracting the mean and get 
\[
  \nabla_\mu (E^\lambda(P^*(\mu, \nu)) = \alpha - \frac{1}{d_1} \left( \sum \alpha_i\right) \mathbf{1}_{d_1}.
\]

Similarly, $\beta$ projected on the mean-zero vectors is the gradient with respect to $\nu$.
This way of obtaining the gradient has been proposed by \cite{Frogner}.

\section{Batch stabilisation}
\label{sec_stable}

For stabilisation we rewrite the iteration \eqref{eq_sk_plain} in log-space as
\begin{equation}
  \label{eq_log_sk}
  \log v^{k+1}_j = \log \nu_j - \logsumexp_{i} (\frac{1}{\lambda} c_{ij} - \log u^{k}_i), \qquad
  \log u^{k+1}_i = \log \mu_i - \logsumexp_{j} (\frac{1}{\lambda} c_{ij} - \log v^{k+1}_j)
\end{equation}
with the log-sum-exp operator $\logsumexp_i x_i = \log \sum_i \exp(x_i)$ that can stably be implemented by extracting the maximum before exponentiation.

When implementing, we would set $\log u_i$ and $\log v_i$ to $-\infty$ if $\mu_i$ or $\nu_i$ is $0$, respectively.

Schmitzer \cite{Schmitzer} proposes to avoid exponentiation and logarithms by splitting $u$ and $\alpha$ and $v$ and $\beta$  and only occasionally absorbing parts into the kernel. While this works well for single $K$, $\mu$, $\nu$, it means that during the iteration, a varying $K$ is used. This does not lend itself to batch computation when one wants to avoid keeping multiple $K$ around. In our experience, the speed of GPGPU computations of one step in the iteration \label{eq_log_sk} tends to be limited by the memory accesses more than the computation.

In \cite{Viehmann}, we provided a batch stabilised version that took the maximum of $\log u$ and $\log v$, but does not have the full stabilisation of the log iteration \eqref{eq_log_sk}.

\section{Implementation for GPGPU}
\label{sec_implementation}

We consider to implement a GPGPU kernel for batches of measures $\mu$, $\nu$ and a single distance matrix.

When using a batch iteration, we need to implement
\[
 \log v_{bj} := \log \nu_{bj} - \logsumexp_{i} (-\frac{1}{\lambda} c_{ij} + \log u_{bi}).
\]

This has two key properties that shape our implementation as an extension to the PyTorch deep learning framework.
\begin{itemize}
\item The overall reduction structure is akin to a matrix multiplication, i.e. memory accesses to $c_{ij}$ and $log u_{bi}$
  to compute the result $\log v_{bj}$, with the additional input $\log \nu$ following the same access pattern as the result. We parallelize in the independent dimensions ($b$ and $j$) and split the reduction over $i$ amongst multiple threads then combine their intermediate results. We have not employed tiling, which is commonly used to speed up the memory accesses for matrix multiplication.

\item In our implementation, the stabilisation of the $\logsumexp$-calculation is carried out in an online fashion, i.e. computing the stabilisation
      and the reduction result in a single pass, similar to the Welford algorithm for the variance.
\end{itemize}

Incidentally, the logarithm of the energy $E^0(P)$ of a minimiser $P$ has a very similar structure, with
\begin{equation}
  \label{eq_E0_calculation}
  E^0(P) = \sum_{ij} p_{ij} c_{ij} = v^T (K \odot P) u = \exp(\logsumexp_j \log v_{j} \logsumexp_i (-\frac{1}{\lambda} c_{ij}+ \log c_{ij} + \log u_i)),
\end{equation}
with $K \odot P$ being the elementwise product, so we can stably compute this loss function with the same GPU kernel as the iteration step.

Here we adopted the point of view that the $K$ and $K \odot P \in \mathbb{R}^{d_1 \times d2}$ can be explicitly computed, but that we would prefer not to realise tensors in  $\mathbb{R}^{batchsize \times d_1 \times d_2}$. For applications with the cost function based on e.g. Euclidean metrics, one might, instead, trade compute for memory and re-create entries of the distance matrix as they are needed.

\section{Practical application in Stochastic Gradient Descent algorithms}

Our goal is to enable the use of the (entropy-regularised) Wasserstein loss for deep learning applications.

As is commonly done, we return $E^0$ as calculated in \eqref{eq_E0_calculation} for the (approximative) minimiser of $E^\lambda$ as the value of our loss function.
The gradient as computed in Section~\ref{sec_backward} is that of $E^\lambda$. Note that the gradient is off for two reasons: First, we use the Lagrange mutliplier for $E^\lambda(P^*)$ as the gradient for $E^0(P^*)$, i.e. we optimize $E^\lambda$ but measure $E^0$. This seems to work reasonably well for many applications and small $\lambda$, but for cases when it does not \cite{Luise} offer an improved gradient. The second source of error ist that the iteration might not have fully converged. Empirically, however, it seems that if we iterate often enough ($1000$ iterations), the gradient is sufficiently good to pass PyTorch's \texttt{gradcheck} tests (we used distribution vectors of length $100$ and $\lambda = 0.001$).

Compared to existing code and libraries, our code combines a stable, memory-efficient logspace implementation that works for batches and uses the Lagrange-Multiplier-based gradient.
In \cite[$\S$ 9.1.3]{PeyreCuturi} advocate the use of automatic differentiation, in the authors' words: \textit{In challenging situations in which the size and the quantity of histograms to be compared
  are large, the computational budget to compute a single Wasserstein distance is usually limited, therefore allowing only for a few Sinkhorn iterations.}
When the histograms do fit the GPU, however, the method of Section~\ref{sec_implementation} seems to achieve a significant speedup over existing implementations, so that in many cases a few tens or even a few hundred iterations seem possible in reasonable time. Also, by not needing to store intermediate results, as relying on autograd implementations of frameworks such as PyTorch, it seems much more memory-efficient to use the Lagrange multiplier. Also, saving the computational cost of backpropagation, which is roughly equivalent to that of the forward pass, allows the number of iterations to be doubled within the same computational budget. As such we disagree with the assessment in \cite{PeyreCuturi}, which is also cited in a recent blog entry \cite{Daza} with implementation.

The latter is particularly important for memory-efficiency because backpropagation though the iteration typically stores intermediate results for each step to facilitate backward computation. This is particularly important because GPU memory is typically an even scarcer resource than computation time in depp learning applications. In our measurement, we achieve a total speedup in forward and backward of 6.5x over \cite{Daza}'s implementation for distributions with mass at 100 points each, even though our choice to not use early stopping causes us compute 3x as many iterations. A significant part of the advantage is that our backward comes at almost negligible computational cost, the remainder from the efficient computational implementation. 

Our code is available at
\url{https://github.com/t-vi/pytorch-tvmisc/blob/master/wasserstein-distance/Pytorch_Wasserstein.ipynb}.

\end{document}